\pdfoutput=1
\documentclass[11pt]{article}

\usepackage{acl}
\usepackage{algorithm}
\usepackage{algorithmic}
\usepackage{times}
\usepackage{latexsym}
\usepackage{enumitem}
\usepackage[T1]{fontenc}
\usepackage{longtable}
\usepackage[utf8]{inputenc}
\usepackage{multirow}
\usepackage{microtype}
\usepackage{graphicx}
\usepackage{inconsolata}
\usepackage{pgfplots}
\usepackage{amssymb}
\usepackage{subfigure}
\usepgfplotslibrary{groupplots}
\usepackage{times}
\usepackage{latexsym}
\usepackage{amsmath}
\usepackage[T1]{fontenc}
\usepackage[utf8]{inputenc}
\usepackage{pgfplots}
\pgfplotsset{compat=1.17}
\usepackage{microtype}
\usepackage{xcolor}
\usepackage{inconsolata}

\title{Optimizing Language Model's Reasoning Abilities with Weak Supervision}

\author{\parbox{0.9\linewidth}{\centering
Yongqi Tong\textsuperscript{1}\thanks{\ \ Equal Constributions.} $\quad$ Sizhe Wang\textsuperscript{2}$^{*}$ $\quad$ Dawei Li\textsuperscript{1}$\quad$ Yifan Wang\textsuperscript{3}$\quad$ Simeng Han\textsuperscript{4}$\quad$ Zi Lin \textsuperscript{1}$\quad$ Chengsong Huang\textsuperscript{5}$\quad$  Jiaxin Huang\textsuperscript{5}$\quad$  Jingbo Shang\textsuperscript{1}}\\
  \textsuperscript{1}University of California, San Diego, \texttt{\{yotong, dal034, lzi, jshang\}@ucsd.edu} \\
  \textsuperscript{2}University of Southern California, \texttt{sizhewan@usc.edu} \\
    \textsuperscript{3}University of Pennsylvania, \texttt{yyifan@seas.upenn.edu} \\
    \textsuperscript{4}Yale University, \texttt{simeng.han@yale.edu} \\
    \textsuperscript{5}Washington University in St. Louis, 
    \texttt{\{chengsong, jiaxinh\}@wustl.edu}\\
}

\begin{document}
\maketitle
\begin{abstract}
While Large Language Models (LLMs) have demonstrated proficiency in handling complex queries, much of the past work has depended on extensively annotated datasets by human experts. 
However, this reliance on fully-supervised annotations poses scalability challenges, particularly as models and data requirements grow. 
To mitigate this, we explore the potential of enhancing LLMs' reasoning abilities with minimal human supervision.  
In this work, we introduce self-reinforcement, which begins with Supervised Fine-Tuning (SFT) of the model using a small collection of annotated questions. 
Then it iteratively improves LLMs by learning from the differences in responses from the SFT and unfinetuned models on unlabeled questions. 
Our approach provides an efficient approach without relying heavily on extensive human-annotated explanations. 
However, current reasoning benchmarks typically only include golden-reference answers or rationales. 
Therefore, we present \textsc{PuzzleBen}, a weakly supervised benchmark that comprises 25,147 complex questions, answers, and human-generated rationales across various domains, such as brainteasers, puzzles, riddles, parajumbles, and critical reasoning tasks. 
A unique aspect of our dataset is the inclusion of 10,000 unannotated questions, enabling us to explore utilizing fewer supersized data to boost LLMs' inference capabilities.  
Our experiments underscore the significance of \textsc{PuzzleBen}, as well as the effectiveness of our methodology as a promising direction in future endeavors.
Our dataset and code will be published soon on \texttt{Anonymity Link}.

\end{abstract}

\section{Introduction}

Large language models~(LLMs)~\citep{brown2020language,zhang2022opt,chowdhery2022palm,touvron2023llama} with Chain-of-Thought (CoT)-based prompting~\cite{wei2022chain,wang2022self,yao2024tree,besta2024graph} have demonstrated strong capabilities across various tasks and applications.
Many previous work to refine LLMs' reasoning abilities have relied on extensive datasets fully annotated by human experts~\citep{longpre2023flan, zhang2022automatic, ranaldi-freitas-2024-aligning, wang2023scott, kim2023cot}. 
This reliance, while beneficial for model training, presents significant scalability challenges.
Although a series of reasoning datasets are published~\citep{amini2019mathqa, cobbe2021training, ling2017program, liu2020logiqa, onoe2021creak, hao2023reasoning, jiang2023brainteaser, joshi2017triviaqa}, the scaling laws indicate that as models grow in size and capabilities, there is an ever-increasing demand for more and updated annotated questions~\citep{hoffmann2022training, sharir2020cost, kaplan2020scaling}, 
which poses a substantial challenge to time and efforts from human supervisors.

\begin{figure*}[t]
    \centering
    \scalebox{0.52}{\includegraphics{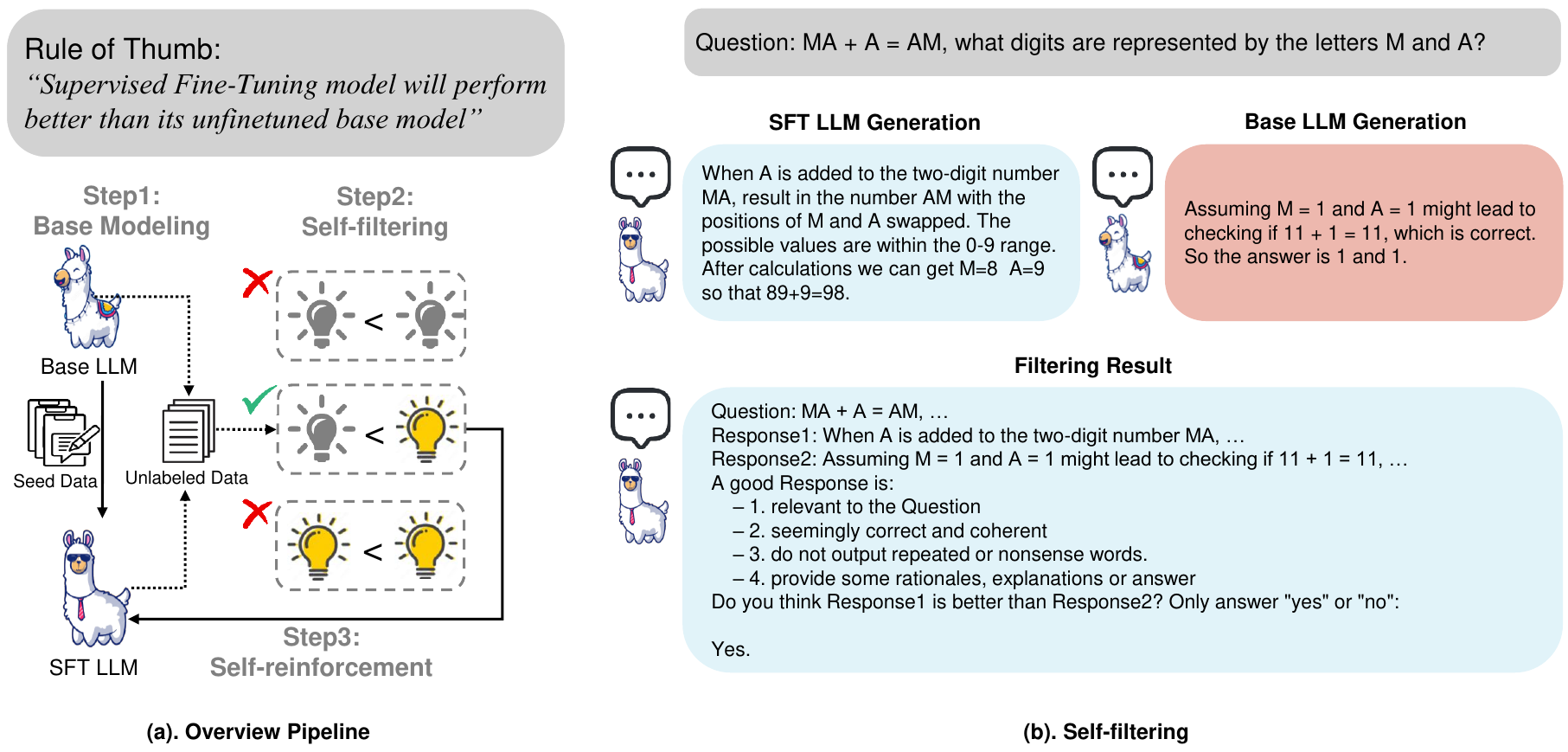} }
    \caption{The overview pipeline of our methods, self-reinforcement and the detailed implementation of self-filtering in our methodology. This is an iterative weak-to-strong learning framework that intends to improve LLMs' reasoning under weak supervision. \textcolor{blue}{Blue content} indicates this response comes from strong models while \textcolor{red}{red content} is from weaker models. 
    % ~\jiaxin{I suggest draw a larger figure for this pipeline to show more details: some small examples of self-filtering, as this has shown to be very effective.}
    }
\end{figure*}

Therefore, an urgent need is to explore how to further advance LLMs' reasoning abilities with fewer human efforts in annotations. 
In this work, we introduce self-reinforcement, a weak-to-strong learning methodology that gained insights from semi-supervised learning. 
This approach is designed for LLMs to iteratively improve their reasoning abilities without reliance on extensive human-labeled rationales. 
We refer to extensive prior research on applying Reinforcement Learning (RL) to preference learning, where a strong learner's thinking process is typically favored over a weaker one~\citep{ziegler2019fine}.
Inspired by this, we intuitively shift the focus from using golden-reference human rationales as absolute positive examples to learning the relative merits between various outputs of the model.

Our methodology unfolds in three phases: initial base modeling, self-filtering, and self-reinforcement. 
Initially, the model undergoes supervised fine-tuning (SFT) using seed dataset to establish a robust foundation for its reasoning capabilities. 
During the self-filtering phase, the model evaluates and eliminates irrelevant or undesired response pairs. 
During the self-reinforcement phase, we hypothesize that the Supervised Fine-Tuned (SFT) model shows better performance compared to its unfinetuned counterpart when addressing unlabeled questions. 
Using Direct Preference Optimization (DPO), we refine the models by learning from the quality differences between their responses to the unlabeled question set. 
This method allows iterative
% \jiaxin{iterative? Not sure what continuous means here}
self-improvement while reducing the reliance on extensively annotated datasets.

Building on our approach, which leverages both supervised and unsupervised learning elements, we recognize the necessity for a tailored dataset.
Therefore, we collect and introduce \textsc{PuzzleBen}, a weakly-supervised benchmark specifically designed to support and validate the effectiveness of weak-to-strong learning paradigms. 
\textsc{PuzzleBen} encompasses a diverse collection of 25,147 labeled questions with answers and meticulously designed human rationale references, as well as 10,000 unlabeled questions.
It consists of various problem types, including brainteasers, puzzles, riddles, parajumbles, and critical reasoning tasks.
The presence of both annotated and unannotated questions within \textsc{PuzzleBen} enables the practical application of our self-reinforcement strategies. 
Additionally, the brainteaser subset in \textsc{PuzzleBen} features with human-labeled difficulty and fun scores, which could be used for further in-depth analysis.

Our experiments highlight the significant impact of human annotated rationales and diverse problem types within \textsc{PuzzleBen}, as well as the efficacy of self-reinforcement in future reasoning work. 

\section{Related Work}
\label{related-work}
\paragraph{LLMs' Reasonings}
CoT~\citep{wei2022chain} equips LLMs with enhanced reasoning capabilities, leading to a series of subsequent studies~\cite{wang2022self,zhou2022least,creswell2022faithful,besta2023graph,li2023making,lightman2023let} that simulate human logical processes. These methods are applied across various reasoning tasks, including commonsense~\cite{geva2021did,ahn2022can}, logical~\cite{pan2023automatically,lei2023boosting}, and mathematical reasoning~\cite{cobbe2021training,hendrycks2021measuring}. 
% Notably, non-linear or lateral reasoning emerges as a distinct category~\citep{hao2023reasoning, jiang2023brainteaser, tong2023eliminating}, characterized by its unique and creative approach that deviates from traditional linear logic to foster innovative problem-solving.

\paragraph{Reinforcement Learning} 
Proximal Policy Optimization (PPO)~\citep{schulman2017proximal} is a key RL technique for aligning models with human preferences~\citep{ouyang2022training}. 
They further lead to the development of Direct Preference Optimization (DPO)~\citep{rafailov2023direct}, which uses the LLM as an implicit reward model. 
Recent efforts are exploring the use of reinforcement learning in tasks that involve reasoning.
For example, \citet{luong2024reft} adopts PPO to differentiate between correct and incorrect reasoning explanations, requiring a large corpus of human-annotated golden references. 
Though this method shows promise, its practical application is uncertain because of inconsistency between rationales and answers generated by LLMs, as mentioned by \citet{luong2024reft, tong2024llms}. 

\paragraph{Self-training and Self-improvement} Many previous works in this direction assign a pseudo label from a learned classifier to further improve the base model~\citep{xie2020self, roychowdhury2019automatic, chen2021semisupervised}. \citet{huang2022large} propose utilizing language models to self-improve without supervised data. 
\citet{chen2024selfplay} employing LLMs from earlier iterations along with human-annotated SFT data to refine the models. 
They contrast data decoded by the models with data supervised by humans and learn from this comparison, which still necessitates considerable human efforts.
Although our work shares similar insights with this direction, we intend to unveil the potential to supervise strong models with a weak model in the field of reasoning.

\paragraph{Weak-to-strong Learning and Generalizations}
\citet{burns2023weak} introduces the potential of leveraging weak model supervision to elicit the full capabilities of much stronger models for superalignment in the future. 
Following this trend, our work tends to explore how to improve LLMs' reasoning abilities under weakly low-resource supervision.
This direction is significant when humans cannot provide large-scale confident answers when the questions become too hard. 

\paragraph{Weakly-supervised Learning}
Many previous works in this field concern about how to benefit from unreliable or noisy labels~\citep{bach2017learning, ratner2017snorkel, guo2018curriculumnet, song2022learning}. 
Semi-supervised learning~\citep{kingma2014semi, laine2016temporal, berthelot2019mixmatch}, when only a subset of labels are available, is closely related to our methodology. 
We fine-tune a base model on a random seed dataset, then iteratively train it on unlabeled data in a semi-supervised manner to progressively improve the initially weak model without full supervision.

\section{Our Methodology: Self-Reinforcement}

In this section, we describe our method to elicit the potential of language models for weak-to-strong generalization in reasoning tasks aimed at minimizing human annotation effort.

Our methodology assumes access to a base language model, a small amount of seed data, and a collection of unlabelled questions. 
The key assumption is that \textbf{Supervised Fine-Tuning (SFT) models will perform better in some questions than its unfinetuned base model within the same training domain}. 
% ~\jiaxin{Currently the two parts (PuzzleBen dataset and Self-Reinforcement technique) are both interesting, but their combination/connection do seem a little loose to me. For example, in PuzzleBen you mention that you have annotated rationales for each question, which is its uniqueness; but for Self-Reinforcement, you use a small amount of seed data and a collection of unlabelled questions. It seems to me the two settings are different (fully supervised vs. weakly-supervised). Reviewers may want to know: (1) where do the unlabelled questions come from, and are they related to PuzzleBen? If the unlabelled questions also come from PuzzleBen, then (2) How well does the Self-Reinforcement learning technique work by training the SFT model with $10\%\rightarrow 20\%\rightarrow\dots\rightarrow 100\%$ of the PuzzleBen training set? (3) How do you choose the iteration number? From Table 4 I can see some results, but it doesn't show what is $t_1$ and $t_2$, so it is hard to interpret the result.}

Our overall pipeline would entail three core steps: 

\begin{itemize}
[itemsep=0mm]
    \item \textit{base modeling:} Access unfinetuned base pretrained model $\pi_0$. Sample a seed data set $ \mathcal{A}^{(0)} = \{(x_\text{g}, r_\text{g}, y_\text{g}) \}$ from the training set in \textsc{PuzzleBen} to optimize an SFT model $\pi_1$ by maximizing $p(r_\text{g},y_\text{g}\mid x_\text{g})$, where $x_\text{g}$ is the sampled question labeled with rationale $r_\text{g}$ and answer $y_\text{g}$.
    % ~\jiaxin{I am a bit confused about this stage. Is $r_0$ generated or directly obtained from PuzzleBen training set?}
    \item \textit{self-filtering:} Sample a set of unlabeled questions $\{x_\text{u}\}$ to generate rationales $r_0 \sim \pi_0(y\mid x_\text{u})$ and $r_1 \sim \pi_1(y\mid x_\text{u})$. We then design a self-filtering prompt to select responses where $r_1$ is preferred over $r_0$ using criteria like relevance and coherence, enhancing the unlabeled dataset with pairs of annotations ${ \mathcal{A}^{(1)} = \{(x_u, r_1, y_1, r_0, y_0) \mid r_1 \succ r_0 \}}$. 
    \item \textit{self-reinforcement:} Then, we apply Differential Performance Optimization (DPO) to learn from the discrepancies between pairs of rationales, further fine-tuning $\pi_1$ on~$\mathcal{A}^{(1)}$ to get~$\pi_2$. 
\end{itemize}

We will describe the procedures of our methodology with more details below. 

\subsection{Step 1: Base Modeling
% \jiaxin{Capitalized ``B'' and ``M'', please do the same for other sections}
}
This initial step involves enhancing the reasoning ability of the unsupervised base model $\pi_0$ by fine-tuning it with a small, high-quality annotated seed data $ \mathcal{A}^{(0)} = \{(x_\text{g}, r_\text{g}, y_\text{g}) \}$,  where $x_\text{g}$ is a sampled question labeled with rationale $r_\text{g}$ and answer $y_\text{g}$.
% ~\jiaxin{changed the expression a little bit}
% , where $r_0$ stands for this query's CoT rationale. 
Given the complexity inherent to our dataset \textsc{PuzzleBen}, each question in the seed data has undergone rigorous examination. 
Any uncertain question will be discussed and vote for the final answer to create rationales. 
% This exhaustive process ensures the golden references, embodying the pinnacle of accuracy and reliability. 
This process is aimed at directly improving the model's basic reasoning ability with the supervised fine-tuning loss function:
% The loss function for a sample can be written
% as in Equation~\ref{eq:sft_loss}. 

\begin{equation}
\small
\mathcal{L}_{\text{SFT}}(\theta) = -\mathbb{E}_{\{(x_\text{g},r_\text{g},y_\text{g})\}\sim \mathcal{A}^{(0)}} \left[ \sum_{t=1}^{|r_\text{g}|} \log (\pi_{\theta}(a_t | s_t)) \right]
\label{eq:sft_loss}
\end{equation}

where \( \theta \) represents the model parameters, 
% \( \mathbb{E}_{\sim\mathcal{D}} \) denotes the expectation over the distribution of $\{(x0,r0,y0)\}$, 
and \( \pi_{\theta}(a_t | s_t) \) is the probability of taking action \( a_t \) at state \( s_t \), given the policy parameterized by \( \theta \). After supervised fine-tuning, we could get $\pi_1 = \pi_\text{SFT}$.
% ~\jiaxin{Should $\sim \mathcal{D}$ be $\{(x_0,r_0,y_0)\}\sim \mathcal{A}_{k}^{(0)}$? Also, should $i=1$ be $t=1$ to keep the notations consistent? Maybe it would be better to replace $L$ with $|r_0|$ or at least explain what is $L$? The notations are not clear and please make all of them consistent throughout the whole paper.}

\subsection{Step2: Self-Filtering}

To select high-quality examples for the next step, we further prompt $\pi_1$ itself to evaluate the response pairs to unlabeled questions generated by itself and $\pi_0$. 
Then we get $r_0 \sim \pi_0(y\mid x_\text{u})$ and $r_1 \sim \pi_1(y\mid x_\text{u})$. 
We attach self-filtering prompting we designed in Table~\ref{tab: self_filtering_prompting}. 
We aim to identify instances where $\pi_1$ outperforms $\pi_0$ based on relevance, coherence, and the presence of detailed rationales.
Only responses where $\pi_1$ demonstrates superior reasoning are retained.

\begin{equation}
\small
{ \mathcal{A}^{(1)} = \{(x_u, r_1, y_1, r_0, y_0) \mid r_1 \succ r_0 \}}
\label{eq:filtering_criterion}
\end{equation}

This selective approach ensures the inclusion of only high-quality rationale annotations in the training process, thereby improving the overall effectiveness of our methods.
% ~\jiaxin{From table 7, it seems the output is a `yes' or `no', so what is the score here?}

\subsection{Step3: Self-Reinforcement}
The third step in our methodology, termed "Self-Reinforcement," employs an innovative approach to further enhance the model's performance. 
This step is based on the assumption that SFT models will exhibit superior rationale-generating capabilities compared to their unfinetuned counterparts within the same training domain. 
This difference in capability is primarily manifested in the quality of rationales produced.

The score \( s_i \) for the output \( (r_i, y_i) \) from \( \pi_i \) and its reference base model \(\pi_{ref}\) is derived as in Equation~\ref{eq:score}. 

\begin{equation}
s_i = \beta\log \frac{P_{\pi_i}(r_i, y_i|x_i)}{P_{\pi_{ref}}(r_i, y_i|x_i)} 
\label{eq:score}
\end{equation}

% We provide new, unlabeled question sets 

% $\{ \mathcal{A}_{k}^{(1)} = (x_1) \}$ 
% to both $M_0$ and $M_1$, allowing each to generate rationales for these questions. 
According to our assumptions, more capable models will obtain higher scores in this phase. 
This output quality discrepancy can be directly learnt with DPO based on the ranking loss in Equation~\ref{eq:self_reinforcement_loss}. 
This enables us to finetune the stronger SFT model $\pi_1$ in a way that systematically amplifies its strengths in rationale generation.

\begin{equation}
\small
L = \sum_{i,j: s_i > s_j} \max(0,s_i - s_j)
\label{eq:self_reinforcement_loss}
\end{equation}

\subsection{Iterative Self-Reinforcement}

Self-reinforcement provides a reasonable approach to continue to refine its own reasoning ability interactively. 
By repeating this process, we enhance the model's understanding and reasoning capabilities to learn from the comparisons between itself and weaker models. 

In the iterative process, we leverage the improved model from the previous iteration, $\pi_1$, and compare its output against the base model, $\pi_0$, to obtain a new model $\pi_2$.
% ~\jiaxin{The notation here is different from previous one where you use $M_1$ and $M_2$.} 
This is formalized as follows:

\begin{equation}
\small
\pi_t = \text{DPO}\left(\pi_{t-1}, \pi_{t-2}\right)
\end{equation}

Where \( \text{DPO}(\cdot) \) represents the Differential Performance Optimization function, taking as input the two models to be compared. 

Notably, our experiments in Section~\ref{sec: experiments} demonstrate that our approach can continually grow with the improvements in the SFT model's capabilities. 
With each iteration of training, the previously "strong" model can serve as the "weaker" model for the next cycle, since the new, stronger model is developed based on the comparison between the two models from the prior round. 

\begin{equation}
\small
L_{\text{iter}} = \sum_{\substack{i,j \\ i \neq j, \\ s_{i}^{t-1} > s_{j}^{t-2}}} \max(0, s_{j}^{t-1} - s_{i}^{t-2})
\end{equation}

Here, \( L_{\text{iter}} \) represents the iterative self-reinforcement learning loss, \( s_{i}^{t-1} \) and \( s_{j}^{t-2} \) represent the scores of the rationales produced by \( \pi_{t-1} \) and \( \pi_{t-2} \) respectively. This iterative process allows the model to improve upon itself, leveraging the comparative strengths of each iteration's outcome.

\section{Data Collection for \textsc{PuzzleBen}}

\begin{figure}[htbp]
    \centering
    \scalebox{0.24}{\includegraphics{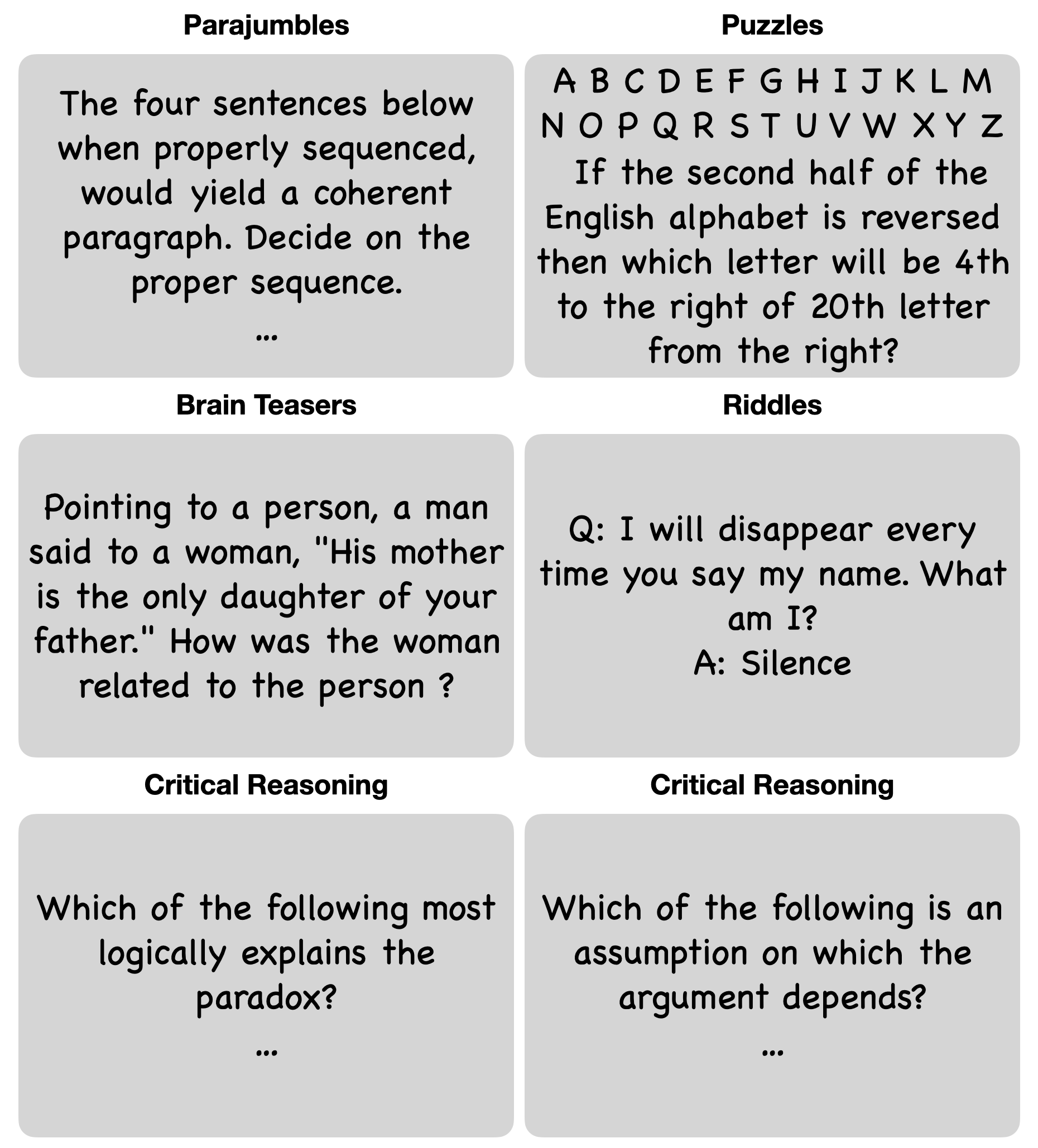}}
    \caption{Question examples from \textsc{PuzzleBen}. The detailed texts are attached in Table~\ref{tab: dataset_example}. 
    % \jiaxin{Can you put this figure to the first page?} 
    }
    \label{fig:benchmarks}
\end{figure}

In this section, we introduce \textsc{PuzzleBen}, a diversified and challenging benchmark with 25,147 annotated questions and 10,000 unannotated queries designed to test and enhance the LLMs' reasoning abilities. 
Our dataset spans multiple domains and question styles, and to illustrate this diversity, we create an overview for questions from  \textsc{PuzzleBen} in Figure~\ref{fig:benchmarks} and include an example question from each category in Table~\ref{tab: dataset_example}. 

Each question in the training set comes with a gold-standard rationale crafted by human experts. 
All the answers and references are well-examined by the websites' users. 
The unlabeled set serves as a special part of \textsc{PuzzleBen} that is pivotal for exploring unsupervised or weakly-supervised learning techniques in the future. 
As for the test set, it has been thoughtfully structured to include options and answers, streamlining the evaluation process for enhanced convenience.
The detailed number of data collected in the three subsets is shown in Table~\ref{tab:subset_size} in Appendix~\ref{sec:appendix}. 

Meanwhile, a distinct section of our \textsc{PuzzleBen} dataset has been enriched with both difficulty and fun scores, informed by user interactions online. 
This feature emerges as a crucial resource for examining the reasoning capabilities of LLMs and their alignment with human supervisory thought processes. 

\label{app:examples}

\subsection{Brainteasers}
The primary intent of collecting brainteasers in \textsc{PuzzleBen} is to promote LLMs' capabilities in tackling problems that require deep thought and creative solutions. 
We systematically collect those questions from a well-designed open-sourced website, Braingle\footnote{\url{https://www.braingle.com/}}. 
 Each question is accompanied by a solution that has garnered widespread acceptance among users, along with a difficulty rating and a human rationale reference. 
 
 A subset of our dataset is distinguished by an additional metric from the website – the success rate of individuals who have attempted. The inclusion of human-assigned difficulty levels and success rates in this subset offers invaluable insights for our subsequent exploration into enhancing the weak-to-strong learning capabilities of LLMs.

\subsection{Riddles}
The primary intent of collecting riddles in \textsc{PuzzleBen} is to compel LLMs to think beyond the immediate context.
A riddle can describe commonsense knowledge in explicit or counterlogical methods~\citep{lin2021riddlesense}.
We collect those well-designed riddles from an open-sourced website famous for stimulating cognitive explosions, ahaPuzzles\footnote{\url{https://www.ahapuzzles.com/}}. 

While ~\citet{lin2021riddlesense} initiated the conversation, our dataset goes a step further by incorporating human rationale, vividly showcasing the intricacies of human thought processes. 
This addition significantly enhances the potential for LLMs to evolve innovatively and critically weak-to-strong generalizations from human's step-by-step reasoning iterations.

\subsection{Puzzles}
Puzzles are designed to challenge our cognitive faculties, forcing us to tap into both learned knowledge and innate logic in real-world problems. 
Unlike riddles, which play on linguistic ambiguities or reconstructing logically coherent narratives, Puzzles hinge on methodical, step-by-step deduction and inference of structured problems. 

We collect puzzles from sawaal\footnote{\url{https://www.sawaal.com/}}, a well-known public website. This aspect is meticulously reviewed and validated by the community, ensuring the dataset serves as a rigorous training ground to promote LLMs from weak and basic capabilities to generalize strong reasoning capabilities.

\subsection{Parajumbles}
Parajumbles involve reordering jumbled sentences into a logical sequence, requiring a deep understanding of the relationships within texts. 
Including parajumbles in our dataset helps transition LLMs from basic learning to advanced modeling, enabling sophisticated logical reasoning.

The inspiration for this task is drawn from two well-known tests - Common Admission Test(CAT)\footnote{\url{https://cdn.digialm.com/EForms/configuredHtml/756/84433/Registration.html}} and Pearson Test of English for Academic(PTE)\footnote{\url{https://www.pearsonpte.com/}}. 
Besides CAT and PTE, we also collect and shuffle those paragraphs from~\cite{misra2022news,harinatha2021evaluating}, two open-sourced news datasets collected from various corpora, such as HuffPost, Business Insider, and CNN.

\subsection{Critical Reasoning}
Critical Reasoning (CR) is essential for evaluating advanced human cognition~\citep{tittle2011critical}. 
Inspired by the reasoning questions from GRE\footnote{\url{https://www.ets.org/gre.html}} and GMAT\footnote{\url{https://www.mba.com/exams/gmat-exam/}}, our CR dataset tests and enhances LLMs' abilities to handle complex logical tasks such as understanding paradoxes, assumptions, and conclusions. This helps LLMs reflect the complex nature of human logic.

While our CR question format is similar to ReClor~\citep{yu2020reclor}, our dataset includes expert rationale from experienced educators and excludes any identical questions found in ReClor, enhancing our benchmark's distinctiveness and educational value.

\subsection{Statistics about \textsc{PuzzleBen}}

\begin{figure}[h]
    \centering
    \scalebox{0.22}{\includegraphics{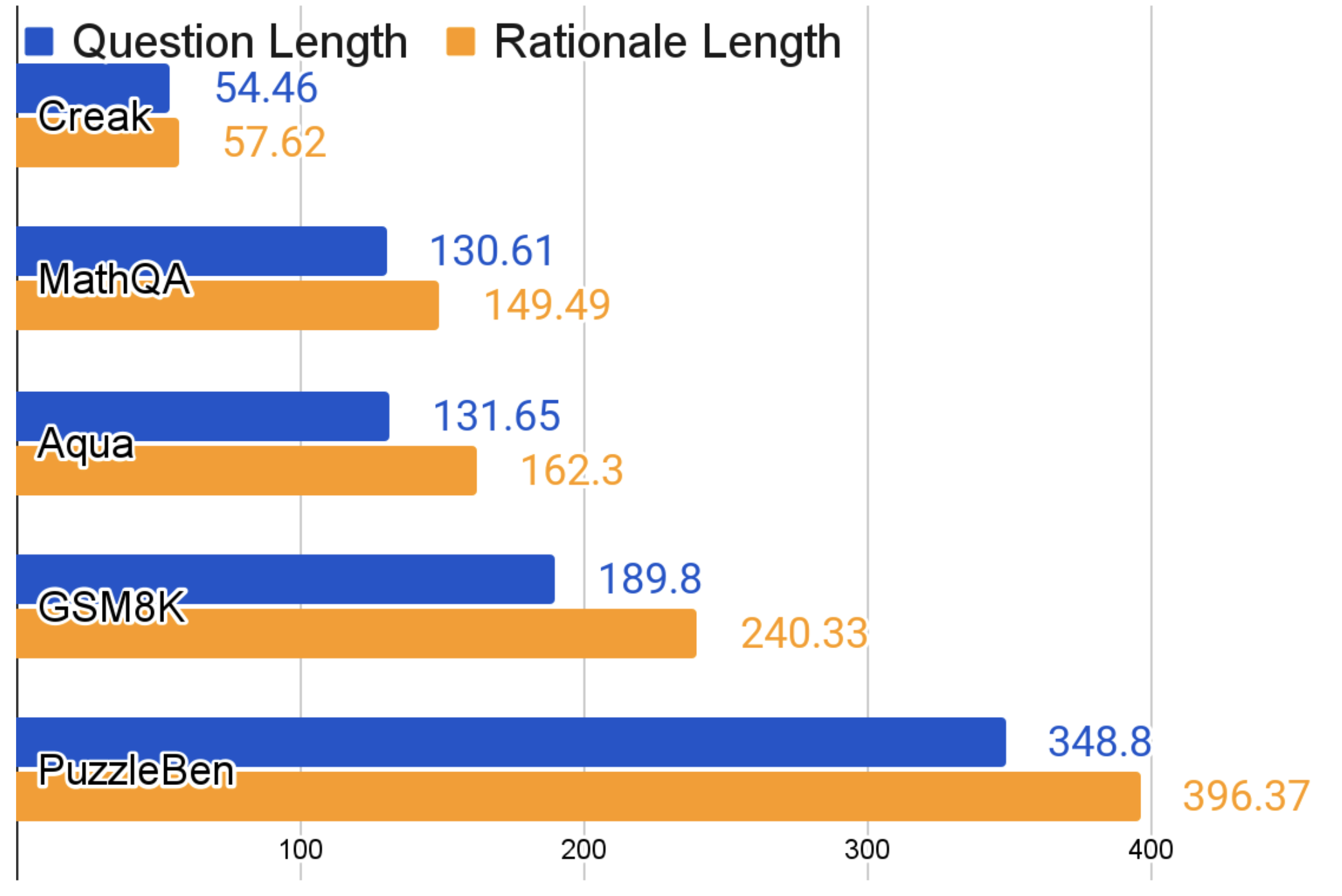}}
    \caption{Average Length of Questions and Rationales designed in \textsc{PuzzleBen} and the other existing benchmarks.
    % \jiaxin{Can you use bar chart for comparison?}
    }
    \label{fig:length}
\end{figure}

In this section, we provide several statistical analyses of our benchmark. 
As we can see in Figure~\ref{fig:length}, \textsc{PuzzleBen} distinguishes itself significantly in terms of the average length of questions and rationales when compared to other existing benchmarks. 
With questions averaging 348.80 characters and rationales at 396.37 characters, PuzzleBen’s content not only exhibits a higher degree of complexity but also provides more elaborate explanations, which further proves \textsc{PuzzleBen}'s uniqueness and necessity to the community. 

A distinctive aspect of our PuzzleBen subset lies in its incorporation of difficulty scores for each brainteaser, derived from the pass rates of online users, offering a directional reflection of our collective grasp on reasoning tasks. The outcomes of our experiments, as detailed in Section~\ref{sec:difficulty_exp}, substantiate the effectiveness and necessity of this feature. This subset promises substantial relevance for future reasoning work, ensuring alignment with human cognitive perceptions from a novel direction.

% \scalebox{0.95}{
%     \begin{tikzpicture}
%     \begin{axis}[
%         title={Average Length of Questions and Rationales \\ designed in \textsc{PuzzleBen} and the other benchmarks},
%         title style={align=center},
%         % ylabel={Average Length},
%         xmin=0, xmax=6,
%         ymin=0, ymax=450,
%         xtick={1,2,3,4,5},
%         xticklabels={Creak,MathQA,Aqua,GSM8K,PuzzleBen},
%         xticklabel style={rotate=45, anchor=east},
%         ytick={0,50,100,150,200,250,300,350,400,450},
%         legend pos=north west,
%         legend entries={Rationale,Question},
%         % axis line style={draw=none},
%         tick style={draw=none},
%     ]
    
%     % Rationale Length with nodes near coords
%     \addplot[
%         color=blue,
%         mark=square,
%         nodes near coords,
%         point meta=explicit symbolic,
%         every node near coord/.append style={font=\small, anchor = south},
%         ]
%         coordinates {
%         (1,57.62) [57.62]
%         (2,149.49) [149.49]
%         (3,162.30) [162.30]
%         (4,240.33) [240.33]
%         (5,396.37) [396.37]
%         };
        
%     % Question Length with nodes near coords
%     \addplot[
%         color=red,
%         mark=triangle,
%         nodes near coords,
%         point meta=explicit symbolic,
%         every node near coord/.append style={font=\small, anchor=north},
%         ]
%         coordinates {
%         (1,54.46) [54.46]
%         (2,130.61) [130.61]
%         (3,131.65) [131.65]
%         (4,189.80) [189.80]
%         (5,348.80) [348.80]
%         };
%     \label{fig:length}
%     \end{axis}
%     \end{tikzpicture}
% }

\section{Baseline Performance on \textsc{PuzzleBen}}

In this section, we evaluate several baseline models' performance an \textsc{PuzzleBen}. 

\subsection{Performance on Five Subtasks}
Table~\ref{tab:PuzzleBen performance} shows standard prompting and zero-shot CoT's performance of GPT4 and PaLM2 on five categories of tasks in \textsc{PuzzleBen}.

As we can see, CoT struggles with the parajumble task. The reason might be that parajumble largely tests concurrent reasoning, where one hypothesizes a sequence and then thinks in reverse to verify its correctness. CoT's step-by-step thinking approach can easily introduce errors at the very beginning of the logic. This limitation underpins the necessity for the \textsc{PuzzleBen} dataset, which aims to enrich future research's landscape by focusing on diverse tasks that challenge current models in various novel ways. 

\begin{table*}[h]
    \centering
    \small
    \begin{tabular}{c|c|ccccc}
    \hline \textbf{Model} & \textbf{Method}     & \textbf{Puzzles} & \textbf{Riddles} & \textbf{Parajumble} & \textbf{CR} & \textbf{Brainteasers}\\\hline
    \multirow{2}{*}{PaLM2} & Standard Prompting~\citep{brown2020language}     & 49.45   & 61.90  & 25.54 &  58.39  & 34.89\\
                          & Zero-Shot CoT~\citep{madaan2023self}  & 53.24   &  63.03 & 20.08 &  51.98 &41.96\\\cline{2-3}\hline

    \multirow{2}{*}{GPT4} & Standard Prompting~\citep{brown2020language}       &  64.37 & 67.70 & 52.17 &  65.32 & 52.58\\
                          & Zero-Shot CoT~\citep{madaan2023self}  & 81.22   &  81.92 & 45.96 & 63.01 &53.53 \\\cline{2-6}
                          \hline
    \end{tabular}
    \caption{PaLM2 and GPT4's accuracy on the five tasks in \textsc{PuzzleBen}. CR stands for critical reasoning subset. }
    \label{tab:PuzzleBen performance}
\end{table*}

\newcommand{\tablefont}{\fontsize{8pt}{6pt}\selectfont}
\begin{table}[h]
    \centering
    \tablefont
    \begin{tabular}{c|ccccc} \hline
      \textbf{Shots}     & \textbf{Puzzles} & \textbf{Riddles} & \textbf{Parajumble} & \textbf{CR}& \textbf{BT} \\\hline
         0   &  81.22   &  81.92 & 45.96 & 63.01 & 53.53\\
         1  & 82.92 & 80.53 & 46.27 & 65.97 &53.02\\
         8  & 84.90  & 85.63 & 51.42 &  68.73 & 55.62\\\hline
    \end{tabular}
    \caption{GPT4's k-shot ICL performance on \textsc{PuzzleBen}. BT stands for Brainteaser tasks.
    % ~\jiaxin{A little confused: The results at ``shots=0'' is the same with \textbf{Zero-Shot CoT} on Puzzles, Riddles, and Parajumble, but is the same with \textbf{Standard Prompting} on CR and BT} 
    }
    \label{tab:huamn_rationale}
\end{table}

\subsection{Utility of Human Rationale Collected in \textsc{PuzzleBen}}

To convince the utility of the human rationales in \textsc{PuzzleBen}, we conduct experiments to utilize those collected rationales both in prompting and fine-tuning directions.  Table~\ref{tab:huamn_rationale} represents the relations between In-Context Learning(ICL) accuracy and k-shot rationale examples.

As the number of shots of the training examples increases, the performance across most tasks seems to improve. 
Specifically, for the Puzzles and Riddles tasks, there's a noticeable increase in performance from the 0-shot to the 8-shot learning. 
The Parajumble and Brainteasers task, though starting with a lower performance score, also shows a similar positive trend. 

The evaluation showcases the utility of human reference in \textsc{PuzzleBen}. It is evident that increasing the number of shots or examples benefits the model's accuracy, especially in tasks like Puzzles, Riddles, Parajumble and Brainteasers. 
This analysis suggests that for tasks demanding a deeper understanding of complex reasoning, a higher number of shots might provide better guidance to the model, leading to improved outcomes.

To further demonstrate the effectiveness of our \textsc{PuzzleBen} dataset, we have conducted a detailed analysis of the effectivenss of collected human rationales in \textsc{PuzzleBen} for SFT.
% ~\jiaxin{Since you mention ``various'', will you finetune the dataset on other backbone models? If not, just say that we analyze the effectivenss of using PuzzleBen for SFT.} 
The results, as shown in Table~\ref{tab:PuzzleBen_testset}, highlights the substantial improvements in LLaMA-13b's performance when finetuned with our dataset.
These improvements underscore the quality and relevance of the training data provided in our \textsc{PuzzleBen}. 
All of those results indicate how well our dataset is suited for enhancing LLMs' complex reasoning capabilities.

% These findings strongly suggest that \textsc{PuzzleBen} serves as a valuable resource for developing and refining AI capabilities in solving puzzles, riddles, and similar tasks that require advanced cognitive skills. The dataset provides a robust foundation for models to learn and adapt to complex problem-solving tasks, thereby proving its effectiveness and utility in the field of artificial intelligence training.

\begin{table}[h]
    \centering
    \small
    \begin{tabular}{c|c|c}
    \hline \textbf{Model}  & \textbf{Method}   & \textbf{Accuracy} \\\hline
    \multirow{2}{*}{LLaMA2-13b} &  - & 10.38 \\
    & after SFT & 36.04 \\\hline
    
    \end{tabular}
    \caption{LLaMA-13b's performance on \textsc{PuzzleBen}'s testset before and after Supervised Finetuning (SFT).
    % ~\jiaxin{The word ``Unfinetune'' seems a bit strange. Maybe just leave it blank, and write ``+SFT'' on the second line?} 
    }
    \label{tab:PuzzleBen_testset}
\end{table}

\subsection{Correlation between Model Performance and Human Difficulty Perception }
\label{sec:difficulty_exp}
\begin{figure}[h]
    \centering
    \scalebox{0.17}{\includegraphics{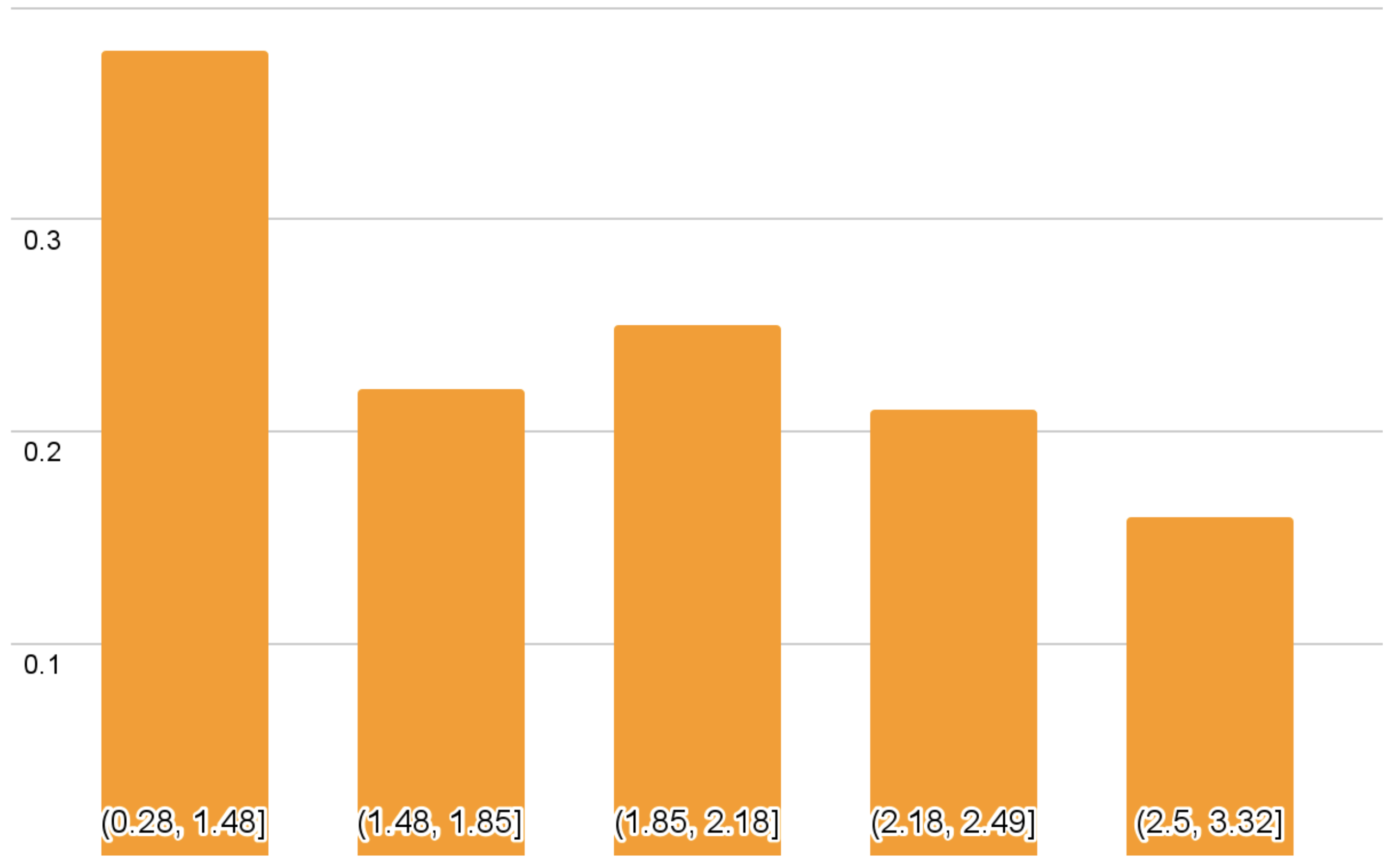}}
    \caption{Accuracy of Llama2-13b across interval-based difficulty score ranges on the subset of \textsc{PuzzleBen}. The difficulty ratings represent the average of all user-assigned scores ranging from 1 to 4, with each category containing an equal number of items.
    % ~\jiaxin{Can you justify here why the last column has a difficulty range larger than previous columns?}
    % ~\jiaxin{I feel the difficulty score here is hard to interpret. Could you add another column for each group to show the human accuracy under each difficulty group? This way we can compare human and Llama2 accuracy in different groups. }
    }
    \label{fig:difficulty_accuracy}
\end{figure}

Our experiments 
Results depicted in Figure~\ref{fig:difficulty_accuracy} illustrate a broad trend where Llama2-13b's accuracy on the PuzzleBen subset wanes as difficulty score intervals rise. 
This pattern shows that the model's challenges generally match the rising difficulty of tasks as humans perceive them, though not perfectly. Our research points to the possibility of improving model performance by tuning it to align more closely with human perceptions of task difficulty, rather than merely matching answers to questions. This approach could enhance the model's understanding of reasoning tasks.

\section{Experiments about Self-Reinforcement}
\label{sec: experiments}

\subsection{Initialization} 
\paragraph{Seed data \& Unlabeled Questions} We randomly select 6400 questions and its rationales from \textsc{PuzzleBen}. 
Considering the difficulty of our dataset, each question and answer has all been fully examined and discussed by annotators. 
We also randomly select 6400 unanswered questions for each iteration. 

\paragraph{Training Details} We choose the pretrained LLaMA2-13b~\citep{touvron2023llama} as our base model. 
Throughout the training, we consistently apply standard hyperparameters: a learning rate of 5e-5, a batch size of 16 instances, and a total of 3 training epochs. Besides, we employ QLoRA~\citep{dettmers2024qlora} with a rank of 16, a LoRA alpha set to 32, and a LoRA dropout rate of 0.05.

% \paragraph{}

\paragraph{Baselines} 
As we discussed in Section~\ref{related-work}, we introduced a novel method to improve LLM reasoning abilities with minimal human effort.  
Self-reinforcement's motivations and settings are different from traditional methods utilizing extensive prompting or heavy fine-tuning. 
Hence, we have few comparable baselines. 
However, a similar approach, ReFT~\citep{luong2024reft}, also uses minimal input and RL to enhance LLMs by learning from model-decoded rationales, specifically by sampling reasoning paths and then creating positive and negative pairs based on the final result. 
Although this method aligns with ours to some extent, it cannot be applied to unformatted human rationale texts or datasets lacking an exact answer.

\subsection{Self-reinforcement Results on \textsc{PuzzleBen}}

\begin{table}[h]
    \centering
    \small
    \begin{tabular}{ccc} \hline
    \textbf{Methods} &
      \textbf{Iterations}   &  \textbf{Accuracy}  \\\hline
         Unfinetune & - & 10.38 \\ 
         SFT & - & 17.33 \\
         ReFT & - & 22.47 \\\hline
        self-reinforcement (ours) & $t_1$ & 28.11 \\
        self-reinforcement (ours) & $t_2$ & 37.82\\
        \hline
    \end{tabular}
    \caption{LLaMA2-13b self-reinforcement and the baselines' results on \textsc{PuzzleBen} with the same labeled seed data set.
    % ~\jiaxin{The `Unfinetune' result here is 10.38, different from 25.99 in Table 3, is there any reason for this?}
    }
    \label{tab:self_reinforcement_res}
\end{table}

Our experimental results on the \textsc{PuzzleBen} dataset using our self-reinforcement approach highlight significant enhancements in model performance. Our method surpassed traditional strategies such as Unfinetuned, SFT, and ReFT, reflecting the efficacy of our iterative, weak-to-strong learning framework. From the base accuracy of 10.38\%, our model's accuracy improved drastically to 37.82\% by the second iteration ($t_2$), underscoring the potential of self-reinforcement in leveraging weak supervision for substantial gains in reasoning tasks.

These findings support the effectiveness of our self-reinforcement methodology in continuously refining the reasoning capabilities of language models under limited supervision. By iterating through cycles of self-filtering and differential performance optimization, our approach not only enhances the quality of rationale generation but also steadily increases the overall model accuracy.

% \subsection{Self-reinforcement Results on Existing Benchmark}

\subsection{Ablation Study}

\begin{table}[h]
    \centering
    \small
    \begin{tabular}{c|c|c} \hline
       \textbf{Iterations}  & \textbf{Methods} & \textbf{Accuracy} \\\hline
       - & SFT & 17.33 \\\hline
       \multirow{2}{*}{$t_1$}  & w/o self-filtering & 18.32\\
     &   w self-filtering & 28.11\\ \hline
             \multirow{2}{*}{$t_2$}  & w/o self-filtering & 18.28\\
     &   w self-filtering & 37.82\\ \hline             
    \end{tabular}
    \caption{Our method's accuracy with and without self-filtering in each iteration. }
    \label{tab:self_filtering_res}
\end{table}

In this ablation study, we further explore self-filtering's potential impacts on our method.
The results in Table~\ref{tab: self_filtering_prompting} distinctly illustrates the crucial role of self-filtering in enhancing the performance of our self-reinforcement methodology. 
By comparing the results of models trained with and without the self-filtering component, it becomes evident that self-filtering significantly boosts accuracy across multiple iterations.

For instance, at iteration $t_1$, the model incorporating self-filtering achieved an accuracy of 28.11\%, which is a substantial increase compared to the 18.32\% accuracy of the model without self-filtering. Similarly, at iteration $t_2$, the gap widened even further, with the self-filtering model reaching an accuracy of 37.82\% compared to 18.28\% for the model without this feature. This clear disparity underscores the effectiveness of self-filtering in refining the dataset and improving the model's reasoning capabilities, thus leading to better performance on complex reasoning tasks.

\section{Conclusions and Future Work}

In this work, we introduce \textsc{PuzzleBen}, a benchmark tailored to augment and assess LLMs' understanding of creative, comprehensive, and non-linear reasoning tasks. 
Each question is designed with high-quality and well-designed rationale reference annotated by human experts. 
In this direction, we propose self-reinforcement, in order to unveil LLMs' weak-to-strong self-learning capabilities in reasoning tasks under weak human supervision.
Our methodology only requires a small annotated dataset compared with previous work. 
To utilize DPO for learning from the quality differences between the rationales decoded by stronger models and those from weaker base models, self-reinforcement provides a possible solution to exploit minimal human supervision effectively.

In future work, we plan to improve the self-reinforcement framework by incorporating dynamic and adaptive self-filtering criteria to enhance the quality of model-decoded data. Furthermore, employing active learning strategies or collaborative human-in-the-loop interventions may help align the models with complex human reasoning techniques and guide the development of LLMs from weak to strong reasoning capabilities. These improvements will aid in creating more autonomous, efficient, and robust reasoning models.

\section*{Limitations}
It is crucial to recognize that the self-reinforcement process could see improvements with further refinements in self-filtering. 
Specifically, choosing more impactful positive and negative pairs can greatly enhance the effectiveness of DPO training.
This approach aligns with the strategy of leveraging highly capable models or human experts for alignment tasks.
% Additionally, incorporating a small proportion of annotated questions into each iteration might enhance the model's learning efficacy and overall performance. These adjustments could lead to more refined and reliable reasoning capabilities in the models developed using our framework. 
Moreover, there remains uncertainty regarding the stability of our model with extensive iterations; specifically, whether the model might experience collapse or increased hallucination phenomena as iterations progress. Introducing a certain proportion of human-annotated data in each iteration could serve as an alignment mechanism, potentially mitigating these issues and ensuring the model remains robust and accurate over long-term training.
% \section*{Acknowledgements}

\bibliography{custom}

\appendix

\section{Appendix}
\label{sec:appendix}

\newcommand{\myfontsize}{\fontsize{8.5pt}{1pt}\selectfont}
\begin{table*}[h]
    \centering
    \myfontsize
    \noindent\rule{\textwidth}{1pt}
    \begin{itemize}[leftmargin=*,label={},topsep=0pt, partopsep=0pt, parsep=0pt, itemsep=0pt]
    \item \textbf{Part 1: Brainteasers} \\
    \begin{itemize}
        \item \textit{Question:} What characteristic do these three 12-digit numbers share with each other, but with no other 12-digit number? \\
    100307124369, 
    111824028801, 
    433800063225.
    \item \textit{Rationale:} 
    \begin{itemize}
        \item They are all square numbers:
        \begin{align*}
        100307124369 &= 316713^2, 
        111824028801 &= 334401^2, 
        433800063225 &= 656635^2.
        \end{align*}
        \item The sum of their digits are square numbers:
        \begin{align*}
        &1+0+0+3+0+7+1+2+4+3+6+9 = 36 = 6^2, 
        &1+1+1+8+2+4+0+2+8+8+0+1 = 36 = 6^2, 
        &4+3+3+8+0+0+0+6+3+2+2+5 = 36 = 6^2.
        \end{align*}
        \item The sum of their digit pairs are square numbers:
        \begin{align*}
        &10+03+07+12+43+69 = 144 = 12^2, 
        &11+18+24+02+88+01 = 144 = 12^2, 
        &43+38+00+06+32+25 = 144 = 12^2.
        \end{align*}
        \item The sum of their digit triplets are square numbers:
        \begin{align*}
        &100+307+124+369 = 900 = 30^2, 
        &111+824+028+801 = 1764 = 42^2, 
        &433+800+063+225 = 1521 = 39^2.
        \end{align*}
        \item The sum of their digit quadruplets are square numbers:
        \begin{align*}
        &1003+0712+4369 = 6084 = 78^2, 
        &1118+2402+8801 = 12321 = 111^2, 
        &4338+0006+3225 = 7569 = 87^2.
        \end{align*}
        \item The sum of their digit sextuplets are square numbers:
        \begin{align*}
        &100307+124369 = 224676 = 474^2, 
        &111824+028801 = 140625 = 375^2, 
        &433800+063225 = 497025 = 705^2.
        \end{align*}
    \end{itemize}
    \item  \textit{Difficulty: 3.23, Fun: 2.45}
    \end{itemize}
        \noindent\rule{\textwidth}{1pt}
    \item \textbf{Part 2: Riddles} \\
    \begin{itemize}
        \item \textit{Question:} What has 13 hearts, but no other organs? \\
        \item \textit{Rationale: } A deck of playing cards consists of 52 cards, divided into four suits: hearts, diamonds, clubs, and spades. Each suit contains one card for each rank from two to ten, plus a jack, queen, king, and ace. This means there are exactly 13 cards in the hearts suit, each metaphorically referred to as having a heart. However, these cards, being inanimate objects, do not possess any other organs, unlike living beings which have a heart along with other organs. This riddle plays on the word hearts as a suit in playing cards and the literal organ, making a deck of playing cards the correct answer since it metaphorically has 13 hearts but lacks any other organs.\\
    \end{itemize}

    \noindent\rule{\textwidth}{1pt}
    \item \textbf{Part 3: Puzzles} \\
    \begin{itemize}
        \item \textit{Question:} 
                A, B, C, D and E are sitting in a row. B is between A and K Who among them is in the middle ?  I. A is left of 13 and right of D.  II.C is at the right end.  [Options] A. If the data in statement I alone are sufficient to answer the question  B. If the data in statement II alone are sufficient answer the question C. If the data either in I or II alone are sufficient to answer the question;  D. If the data in both the statements together are needed. \\
        \item \textit{Rationale: } Clearly, we have the order : A. a E. From I, we have the order : D, A, B. E. From II, we get the complete sequence as D, A, B. E, C. Clearly. B is in the middle. So, both I and II are required. 
    \end{itemize}
    \noindent\rule{\textwidth}{1pt}
    \item \textbf{Part 4: Critical Reasoning} \\
    \begin{itemize}
        \item \textit{Question:} 
               In the shallow end of Lake Tomwa, there are remains of numerous Jeffery pine trees that grew there during a lengthy drought. Researchers had believed that this drought lasted at least 150 years, but carbon dating reveals that pines were growing in the lake bed for only 120 years, from 1200 until 1320. Since the Jeffrey pines, which cannot survive in water, must have died at the end of the drought, the dating shows that the drought lasted less than 150 years.  \textbf{The argument given relies on which of the following as an assumption?} [Options] A.  No other species of tree started growing in the bed of Lake Tomwa after 1200. B.  No tree remains of any kind are present at the bottom of deeper parts of Lake Tomwa. C. There was at least one tree in the lake bed that was alive for the entire period from 1200 to 1320. D.  There has not been a more recent drought that caused a drying up of the shallow end of the lake. E. The shallow end of the lake had been dry for less than 30 years by the time Jeffrey pines started growing in the lake bed. \\
        \item \textit{Rationale: } The reasoning process in this article can be summarized as follows: (1) Pine trees cannot survive in water (they can only survive during dry periods) → after the dry period ends, J pine trees will inevitably die; (2) J pine trees only lived for 120 years: (1)+(2) → the duration of the drought was less than 150 years. The problem with this reasoning process is that it cannot determine when the drought began, as the drought could have started well before the J pine trees began to grow. Option A is incorrect because whether other species of trees began to grow 1200 years later does not affect the inference in the text, as the dating method mentioned is specific to J pine trees and is not influenced by other species of trees. Even if other water-resistant species of trees survived, it is irrelevant to the discussion at hand. Option B is incorrect, as whether trees existed at the deeper bottom of the lake does not affect the inference in the text. The depth of the lakebed where trees grew at most could only indicate the extent of the drought, not the existence of the drought itself. Option C is incorrect because whether any trees lived through the entire 120 years does not affect the inference in the text, as the dating method mentioned has already proven that J pine trees grew from 1200 to 1320. Even if each tree lived only one year, it does not affect the deduction that "J pine trees survived between 1200 and 1320." Option D is incorrect because whether a drought occurred again later does not affect the inference in the text, as whether there was a drought later is irrelevant to the study of this period. Additionally, the dating method has already proven that pine trees only survived during the consecutive 120 years between 1200 and 1320, which indicates that the specific drought period mentioned ended in 1320. Option E is correct because the text does not provide evidence on when the drought began. If the drought had already lasted for more than 30 years by the time J pine trees began to grow, then adding the 120 years of J pine trees' growth period, the total duration of the drought would exceed 150 years, contradicting the conclusion in the text.
    \end{itemize}
\noindent\rule{\textwidth}{1pt}
    \item \textbf{Part 5: Parajumble} \\
    \begin{itemize}
        \item \textit{Question:} 
    Reorder the following sentences to form a coherent paragraph. Sentence A) For example, if I am a group member, I can choose group -sending. Sentence B) About what an email list is. Sentence C) What the use of email list is. You can arrange contacts into a particular group in the email list. Sentence D) Further explanation for the example. No new words, and very easy. 
    \item \textit{Rationale: }
    To solve this, we shall analyze the given sentences closely to understand their logical and thematic connections. Sentence B serves as a general introduction by talking about what an email list is. It sets the stage for further discussion on the specifics of an email list, making it the natural starting point. Following the introduction of the email list, Sentence C delves into What the use of email list is by explaining that You can arrange contacts into a particular group in the email list. This explanation directly builds upon the introductory concept provided in sentence B, expanding the reader\'s understanding of the functionality and purpose of an email list. Sentence A presents a specific example For example, if I am a group member, I can choose group-sending. This sentence illustrates a practical application of the concept introduced in sentences B and C, showing how an individual might utilize the email list\'s functionality. Finally, Sentence D offers Further explanation for the example. No new words, and very easy. Since it aims to elaborate on the example given in sentence A, it logically follows that sentence, rounding off the explanation and providing clarity. Thus, the coherent sequence is B (introduction to the topic), followed by C (explanation of usage), leading into A (specific example of usage), and concluded with D (further elucidation of the example). Therefore, the correct order is BCAD, creating a logical flow from a general introduction to a specific example and its explanation.
    \end{itemize}
\end{itemize}
% \noindent\rule{\textwidth}{1pt}
    \noindent\rule{\textwidth}{1pt}
\caption{Examples collected from our \textsc{PuzzleBen}. }
\label{tab: dataset_example}
\end{table*}

\begin{table*}
\noindent\rule{\textwidth}{1pt}
\begin{itemize} [itemsep=0mm]
    \item Question: \{\}
    \item Response1: \{\}
    \item Response2: \{\}
    \item A good Response is: 
    \begin{itemize} [itemsep=0mm]
        \item 1. relevant to the Question
        \item 2. seemingly correct and coherent
        \item 3. do not output repeated or nonsense words.
        \item 4. provide some rationales, explanations or answer
    \end{itemize}
        \item Do you think Response1 is better than Response2? Only answer "yes" or "no":

\end{itemize}
\noindent\rule{\textwidth}{1pt}
\caption{Prompting we designed in the stage of self-filtering. Response1 is generated from $M_1$ while Response2 is from $M_0$. We filter out the samples which Response1 is obviously worse than Response0. }
\label{tab: self_filtering_prompting}
\end{table*}

\begin{table*}
    \centering
    \begin{tabular}{cc} \hline
    \textbf{Subset} &
      \textbf{Size}     \\\hline
         Annotated Trainset & 22,528 \\ 
         Unannotated Question Set & 10,000 \\
         Testset & 2,618 \\\hline
    \end{tabular}
    \caption{Detailed Subset's Size in \textsc{PuzzleBen}.}
    \label{tab:subset_size}
    
\end{table*}

\end{document}